\documentclass[10pt,twocolumn,letterpaper]{article}

\usepackage[pagenumbers]{cvpr} 

\definecolor{cvprblue}{rgb}{0.21,0.49,0.74}
\usepackage[pagebackref,breaklinks,colorlinks,allcolors=cvprblue]{hyperref}
\usepackage{graphicx}
\usepackage{amsmath}
\usepackage{amssymb}
\usepackage{colortbl}
\usepackage{booktabs}
\usepackage{amsmath,amsthm,amssymb,amsfonts}
\usepackage[ruled]{algorithm2e} 
\usepackage{enumitem} 
\usepackage{caption}
\usepackage{pifont}
\usepackage{multirow}
\usepackage{enumerate}
\usepackage{color}
\usepackage{lineno}
\usepackage[accsupp]{axessibility}

\newcommand{\av}[0]{\ensuremath{\boldsymbol{a}} }

\newcommand{\dv}[0]{\ensuremath{\boldsymbol{d}} }

\newcommand{\xv}[0]{\ensuremath{\boldsymbol{x}} }

\def\gg{\textcolor{gray}}

\usepackage[capitalize]{cleveref}
\crefname{section}{Sec.}{Secs.}
\Crefname{section}{Section}{Sections}
\Crefname{table}{Table}{Tables}
\crefname{table}{Tab.}{Tabs.}

\def\paperID{6421} 
\def\confName{CVPR}
\def\confYear{2025}

\title{Explaining Domain Shifts in Language: Concept erasing for \\ Interpretable Image Classification}

\author{Zequn Zeng$^1$\thanks{Equal contribution. \hspace{4mm}  \textdagger Corresponding authors},  Yudi Su$^1$\footnotemark[1], Jianqiao Sun$^1$, Tiansheng Wen$^1$, \\
Hao Zhang$^1$, Zhengjue Wang$^2$, Bo Chen$^1$\footnotemark[2], Hongwei Liu$^1$\footnotemark[2] and Jiawei Ma$^3$\\
$^1$National Key Laboratory of Radar Signal Processing, Xidian University, Xi’an, 710071, China.\\
$^2$State Key Laboratory of Integrated Service Networks, Xidian University, Xi’an, 710071, China.\\
$^3$City University of Hong Kong, Hong Kong SAR.\\
{\tt\small zqzeng\_1@stu.xidian.edu.cn, bchen@mail.xidian.edu.cn }
}

\begin{document}
\maketitle
\begin{abstract}
Concept-based models can map black-box representations to human-understandable concepts, which makes the decision-making process more transparent and then allows users to understand the reason behind predictions.
However, domain-specific concepts often impact the final predictions, which subsequently undermine the model generalization capabilities, and prevent the model from being used in high-stake applications.
In this paper, we propose a novel \textbf{Lan}guage-guided \textbf{C}oncept-\textbf{E}rasing (LanCE) framework. 
In particular, we empirically demonstrate that pre-trained vision-language models (VLMs) can approximate distinct visual domain shifts via domain descriptors while prompting large Language Models (LLMs) can easily simulate a wide range of descriptors of unseen visual domains.
Then, we introduce a novel plug-in domain descriptor orthogonality (DDO) regularizer to mitigate the impact of these domain-specific concepts on the final predictions. 
Notably, the DDO regularizer is agnostic to the design of concept-based models and we integrate it into several prevailing models. Through evaluation of domain generalization on four standard benchmarks and three newly introduced benchmarks, we demonstrate that DDO can significantly improve the out-of-distribution (OOD) generalization over the previous state-of-the-art concept-based models. Our code is available at
\href{https://github.com/joeyz0z/LanCE}{https://github.com/joeyz0z/LanCE}.

\end{abstract}

\section{Introduction}
\label{introduction}

\begin{figure}[!t]
  \centering
  \subfloat[Concept-based predictions on three domains. ]{\includegraphics[width=0.8\linewidth]{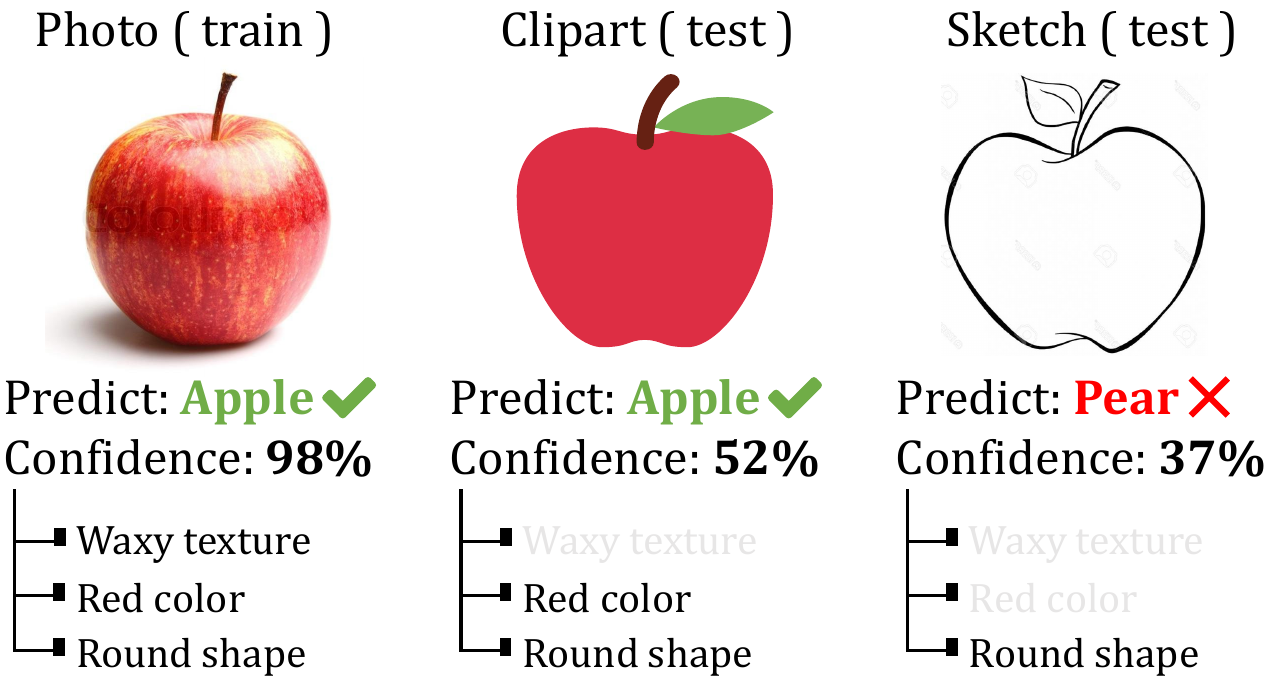}\label{fig1a}} 
  \quad
  \subfloat[Distribution shifts of domain-specific and domain-shared concepts.]{\includegraphics[width=1.0\linewidth]{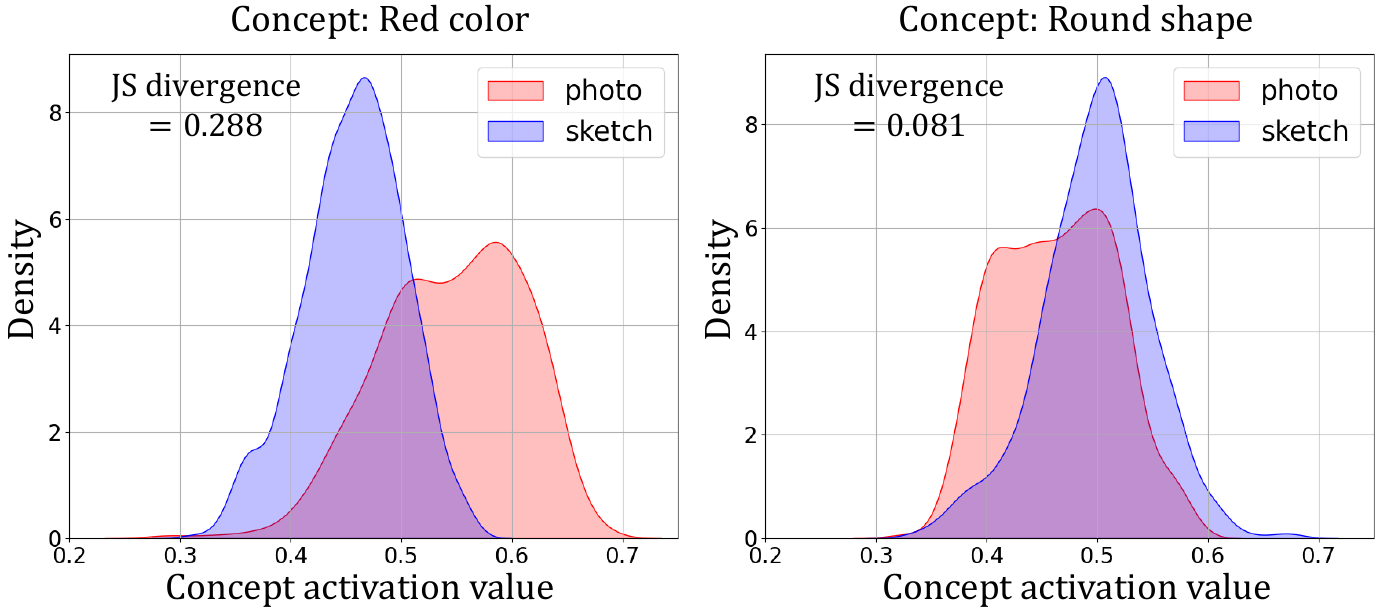}\label{fig1b}} 
    \quad
\vspace{-3mm}
  \caption{Domain shifts in concept space. (a) Given the images of an apple in different visual domains, the prediction confidence of a concept-based model, trained on the photo domain, degrades due to the missing of concepts. 
  (b) Distribution comparison of concept activation value (image-concept similarity computed via CLIP~\cite{clip}) between photo domain and sketch domain, for two concepts, \ie, ``red color'' and ``round shape'', respectively. JS divergence indicates the distance between two distributions and tends to be larger for domain-specific concepts ($\eg$ ``red color'').}
  \label{fig1}
\end{figure}

Concept-based models~\cite{cbm,conceptembeddings,post-hoc-cbm} are prominent approaches for achieving model interpretability, which leverage human-understandable concepts to explain the black-box image representation. 
Specifically, these models first map the image feature to a concept activation vector in a concept space (each dimension corresponds to an interpretable concept) and then use the concept activation vector to predict the final output.
Recently, equipped with pre-trained vision-language models (VLMs)~\cite{clip,align}, concept-based models~\cite{labo,post-hoc-cbm} can obtain the concept activation value by calculating the similarity between image embeddings and textual concept embeddings.

Nevertheless, current concept-based models
still struggle to handle domain shifts encountered during inference. 
The distinct distribution of visual concepts across different domains presents a significant generalization challenge for these models.
As shown in Fig.~\ref{fig1a}, the concept-based models trained on the photographic images usually excel at associating the discriminative visual clues -- such as red color, waxy texture, and round shape -- with specific classes like ``apple'', akin to human beings. 
However, these models experience substantial performance degradation when applied to images in unseen visual domains, such as clipart or sketches, due to the absence of domain-specific texture and color concepts. 
These domain-specific concepts ($\eg$ ``red color'') generally exhibit a larger distributional difference between visual domains than domain-shared concepts ($\eg$ ``round shape''), as illustrated in Fig.~\ref{fig1b}.
Consequently, the concept-based models trained on a single domain are inevitably biased to associate those domain-specific concepts with the final predictions, thereby limiting their out-of-distribution (OOD) generalization capabilities.

To mitigate the issue of domain shifts in the concept space, a straightforward solution is to involve human experts in filtering out the domain-specific concepts and retaining only domain-shared concepts for classification tasks, which is nevertheless labor-intensive.
Another possible solution is to diversify the training domain~\cite{chen2023meta,diversify} to mitigate the impact of the domain-specific concepts. 
However, it is impractical to cover all possible unseen domains during training exhaustively.
Additionally, current domain generalization approaches~\cite{zhou2022domain} can not be directly applied to concept-based models due to the lack of interpretability of semantics of the learned representation~\cite{interpretableDG}.
Therefore, automating the identification and elimination of domain-specific concepts remains an under-explored area.

\begin{figure*}[!tb]
	\centering 
	\includegraphics[width=1.0\textwidth]{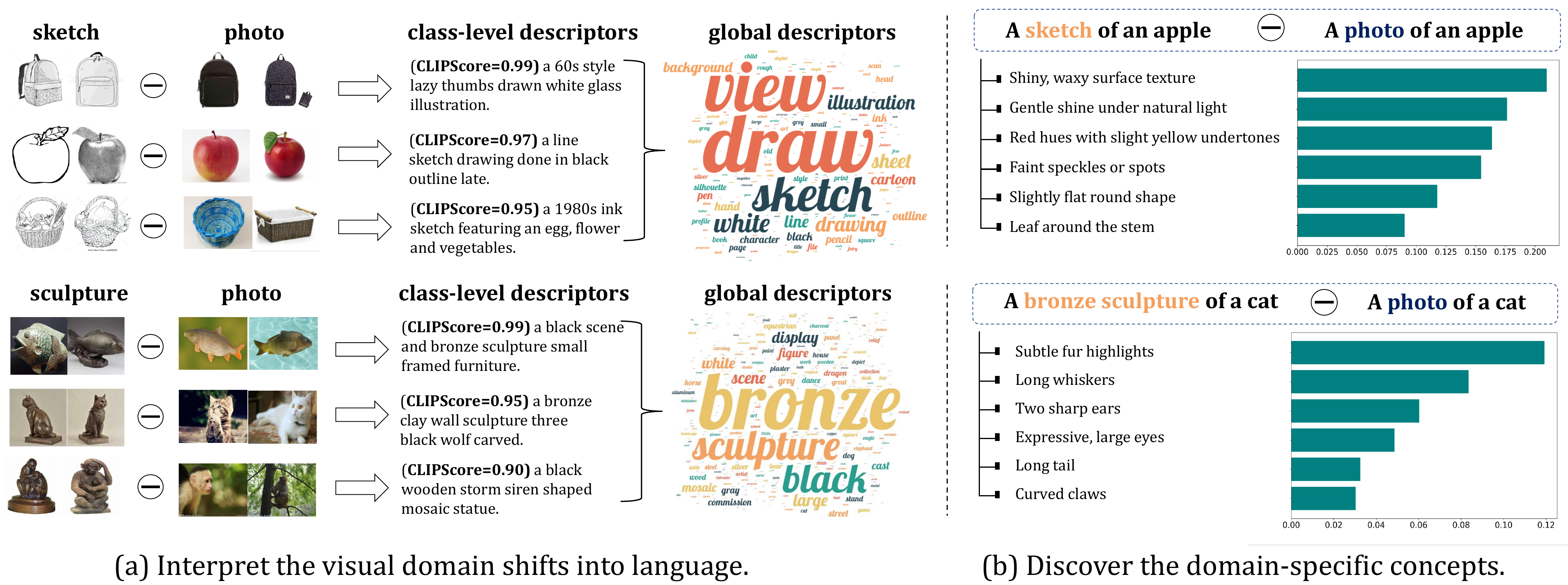}
        \vspace{-6mm}
	\caption{(a) We empirically demonstrate that the visual domain shift can be interpreted in language. For each class, we obtain the caption from the embedding difference of images from two domains. Then, by aggregating the captions across the classes, the keywords regarding the domain shift can be highlighted. At the same time, (b) the domain-specific concepts can be discovered from language descriptions of the different domains. In detail, they have higher similarities with the difference of domain-related class descriptions, \ie, textural domain shift, in the CLIP embedding space. More analyses are shown in the Appendix \ref{appendix:empirical}.}\label{empiricalObservation}
\end{figure*}

In this paper, we propose a \textbf{Lan}guage-guided \textbf{C}oncept-\textbf{E}rasing framework, namely \textbf{LanCE}, to enhance the OOD generalization capabilities of current concept-based models. 
Specifically, we empirically demonstrate that pre-trained VLMs, such as CLIP, can interpret the domain shifts through language, $\ie$, domain shifts can be approximated by a set of domain descriptors generated by large language models (LLMs).
To alleviate the biased association between final predictions and domain-specific concepts, we introduce a plug-in domain descriptor orthogonality (DDO) loss that reduces the fluctuation of concept activation caused by these language-guided domain shifts.
We evaluate our method on seven benchmarks, including four common benchmarks (CUB-Painting~\cite{cub-painting}, PACS~\cite{pacs}, OfficeHome~\cite{officehome} and DomainNet~\cite{domainNet}), and three new proposed benchmarks (AwA2-clipart, LADA-Sculpture and LADV-3D) based on
several existing visual classification datasets with concept annotations~\cite{awa2,lad}. 
These newly-collected datasets are gathered from the web and manually filtered. 
Extensive experiments demonstrate that the proposed approach can preserve the In-distribution (ID) classification capabilities and significantly improve the OOD generalization capabilities of prevailing concept-based models.
Our contributions can be summarized as follows:
\begin{itemize}
\item We provide a concept-level explanation for the domain shifts and empirically reveal that VLMs can represent visual domain shifts into language space.
\item Based on this observation, we propose a plug-in domain descriptor orthogonality regularizer to learn a domain-shared concept-based model without changing the model architecture and increasing training data. 
\item We introduce three new concept-based OOD generalization benchmarks including more challenging scenarios such as natural photos$\rightarrow$3D-renders and real animals$\rightarrow$scluptures. Extensive experiments on four common and three new benchmarks demonstrate that our proposed method significantly improves the OOD generalization capabilities of current concept-based models.


\end{itemize}

\section{Related Work}
\textbf{Concept-based models}
are one of the mainstream interpretable approaches that aim to understand models by associating black box features with meaningful concepts.
Current concept-based models can broadly be divided into two categories, concept bottleneck models (CBMs)~\cite{cbm} which map the image features into an intermediate concept bottleneck layer whose neurons indicate concept activation values of pre-defined concepts, and concept activation vectors (CAVs)~\cite{cav} which represent concepts as normal vectors of decision boundaries that distinguish positive and negative samples of a concept. Vanilla CBM requires fine-grained and precise concept annotation on the concept bottleneck layer and CAVs need positive and negative samples of each concept to learn meaningful concept representations. 
Many recent approaches are proposed to tackle the above concerns from two perspectives. On the one hand, many methods generate the concepts derived from ConceptNet~\cite{post-hoc-cbm}, LLMs~\cite{labo,label-free} or multimodal datasets~\cite{shang2024incremental,cdl} to automate the construction of the concept bottleneck layer. On the other hand, to remove the reliance on human-annotated concept labels, these methods deploy pre-trained VLMs such as CLIP to map concepts to text embeddings and compute image-concept similarity to serve as concept activation annotations.
However, concepts of the above approaches are typically domain-sensitive and the corresponding concept activations have large distributional differences across various domains. They often fail to handle domain shifts when applied to unseen domains.

\noindent \textbf{Domain adaptation \& generalization} aims to handle the domain shifts between training and test data. In contrast to traditional deep learning based on in-distribution assumption, domain adaptation methods~\cite{domainAdaption,domainAdaption1} target to improve the performance on the target domain when only a few samples~\cite{semidomainAdaption} or unlabeled data~\cite{Unsuperviseddomain} from target domains are available. Instead, domain generalization methods~\cite{zhou2022domain} focused on more difficult scenarios in which no target domain data are available during the training phase. 
The setting of this paper belongs to the single domain generalization scenario, $\ie$ only one single domain is available during training~\cite{wang2021learning,single1,single2}. While the domain generalization area is well studied, there is still researcher doubt about the interpretability of domain generalization methods~\cite{interpretableDG}. Although disentanglement-based domain generalization methods~\cite{csd, distengleDG} decompose a feature into domain-shared and domain-specific parts, there is still a lack of deep understanding of the semantics of the learned features in domain generalization models. Our proposed method provides a concept-level explanation of domain shifts and takes steps to interpretable domain generation.

\noindent \textbf{Vision-Language Pretraining} aims to bridge the gap between image and text representations.
Compared with vision-only pretraining methods~\cite{mode,moco,mae}, vision-language pertaining like CLIP~\cite{clip} achieved remarkable success and showed superior performance on some high-level vision understanding tasks~\cite{zeng2023conzic,meacap,luo2021clip4clip} and multi-modality tasks~\cite{cliptask,hicescore,wen2025beyond}. 
Our empirical findings further validate that CLIP can interpret a variety of domain shifts into a set of domain descriptors.


\section{Empirical Observations}
\label{empirical}
This section demonstrates that the pre-trained VLMs can approximate the visual domain shifts via domain descriptors. Meanwhile, these language-guided domain shifts can effectively distinguish the domain-specific concepts as illustrated in Fig.~\ref{empiricalObservation}. 
It serves as the key insight for our proposed LanCE method.

\noindent\textbf{Image-text alignment.} CLIP trained on multi-modal contrastive learning, can effectively bridge the gap between image representation and text representation. 
The alignment facilitates a new approach to manipulate images using language, applicable for various vision tasks, $\eg$ image editing~\cite{styleCLIP,prompt2prompt}, style transfer~\cite{clipstyle,stylediffusion}, data augmentation~\cite{lads,widin,diversify}. 
The CLIP image-text similarity can be computed as:
\begin{equation}
\label{eq: clip}
\mathcal{L}_\text{CLIP}(I,T) = \text{sim}(E_I(I),E_T(T))
\end{equation}
where $E_I$ and $E_T$ mean the image encoder and text encoder of CLIP, respectively. $\text{sim}$ indicate the cosine similarity.

\begin{figure*}[!tb]
	\centering 
	\includegraphics[width=1.0\textwidth]{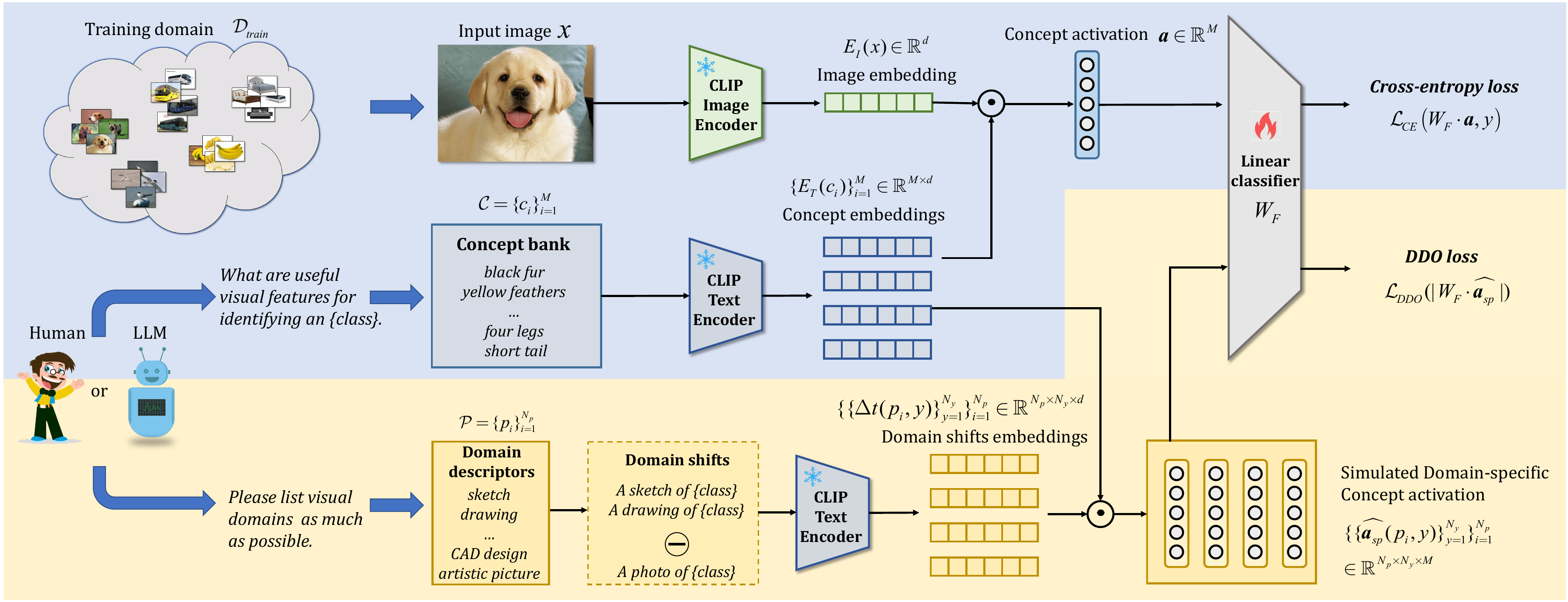}
        \vspace{-5mm}
	\caption{Overview of the LanCE. \textcolor[RGB]{50,50,243}{Blue part} is the data flow of vanilla CLIP-CBMs (Sec.~\ref{Problem formulation}).  To provide concept-level explanations, we first construct a human-written or LLMs-generated concept set $\mathcal{C}$ and extract the concept embeddings via the frozen CLIP text encoder. Given an image, we can extract the image embeddings via the frozen CLIP image encoder. The concept activations are the cosine similarity between image embeddings and concept embeddings. A learnable linear layer $W_F$ is fitted on top of the concept activation vector and is responsible for predicting the final class and is optimized via cross-entropy loss. \textcolor[RGB]{255,192,0}{Yellow part} is the data flow of our proposed DDO regularizer (Sec.~\ref{ddoloss}). Similarly, we first construct a domain descriptor set (Sec.~\ref{Generate domain descriptors}) to obtain the language-guided domain shifts and then simulate the domain-specific concept activations. To erase the effect of domain-specific concepts, the DDO regularizer encourages the orthogonality between the class-concept correlation matrix $W_F$ ($\ie$ the final linear weight) and domain-specific concept activation $\widehat{\av_{\text{sp}}}$.  } 
 \label{overview}
\end{figure*}

\noindent\textbf{Interpreting the visual domain shifts.}  
Inspired by previous works~\cite{styleCLIP,lads}, we assume that the visual domain shifts and descriptions of domain changes are aligned within the CLIP embedding space.
To validate this hypothesis, we employ a CLIP-based zero-shot captioner~\cite{zeng2023conzic} to translate the visual embedding differences into language, as illustrated in Fig.~\ref{empiricalObservation}(a). Let $\{I_c^{src}\}$ and $\{I_c^{tgt}\}$ denote the images of the same class $c$ from the source domain and the target domain, respectively.
The class-level visual domain gap $\Delta \dv_c$ and corresponding description $t_c$ can be generated by solving the following maximization problem:
\begin{align}
\label{eq: clip}
\max_{t_c} &\mathcal{L}_{\text{CLIP}}(\Delta \dv_c, t_c), \\
\Delta \dv_c =&\mathbb{E}(E_I(I_{c}^{tgt})) - \mathbb{E}( E_I(I_{c}^{src})),
\end{align}

As shown in Fig.~\ref{empiricalObservation}(a), we produce high-matching (CLIPScore $>$ 0.9) descriptions $t_c$ that effectively interpret the class-level visual domain gap $\Delta \dv_c$ between sketch images, sculpture images and photographic images. 
We observe that these class-level domain descriptors are typically descriptions of the style difference of the two domains, regarding class-level domain descriptors. 
To further get the global difference between the two domains,  we aggregate all class-level domain descriptors and transfer them into word cloud format, where the word size is proportional to the word frequency in the generated domain description corpus.
As a result, we empirically find that global visual domain descriptors $\Delta \dv$ primarily consist of style words such as ``sketch and sculpture'' and background words $\eg$ ``white''. 


\noindent\textbf{Discover the domain-specific concepts.}  
Since we have revealed that visual domain shifts and descriptions of domain changes are aligned within the CLIP embedding space, we can synthesize the visual domain shifts via domain descriptors.
Subsequently, we can leverage these language-guided domain shifts to discover domain-specific concepts.
Specifically, we compute the similarity between textual domain shift embeddings ($\eg$ ``A sketch of an apple'' - ``A photo of an apple'') and all candidate concept embeddings extracted by CLIP text encoder.
The results in Fig.~\ref{empiricalObservation}(b) show that domain-specific concepts like ``shiny, waxy surface texture'' generally exhibit a higher similarity score due to their substantial distribution shift between those two domains while those domain-shared concepts such as ``Leaf around the stem'' have relatively lower similarity score.
Consequently, we can distinguish the domain-specific concepts via language-guided domain shifts.

\section{Method}
To improve the generalization capabilities of prevailing concept-based models, as shown in Fig.~\ref{overview}, we propose a novel \textbf{LanCE} framework. 
In this section, we first formulate the problem and introduce Concept bottleneck models (CBMs), which can effectively map the black-box visual features to human-understandable concept space to interpret the final outputs (Sec.~\ref{Problem formulation}). However, domain-specific concepts will undermine the generalization capabilities.
Then, to alleviate the biased association between domain-specific concepts and class predictions within the learnable linear classifier, we propose a domain descriptor orthogonality loss to improve the OOD classification performances of concept-based models (Sec.~\ref{ddoloss}).
Furthermore, considering all possible visual domains, we prompt LLMs to generate a bunch of domain descriptors to simulate numerous unseen visual domains (Sec.~\ref{Generate domain descriptors}).

\subsection{Problem Formulation}
\label{Problem formulation}
Given a training domain data $\mathcal{D}_{train}=\{(\xv_i,y_i)\}$, where $\xv_i\in\mathcal{X}$ are training images and $y_i\in\mathcal{Y}$ are corresponding labels,  
the concept-based models are trained only on $\mathcal{D}_{train}$, however evaluated on both $\mathcal{D}_{train}$ and $k$ unseen domains $\{\mathcal{D}_{unseen}^i\}_{i=1}^k$. 
To provide a concept-level explanation, we should construct a concept set $\mathcal{C}=\{c_i\}_{i=1}^M$ including $M$ interpretable concepts written by humans or LLMs. To evaluate the effectiveness of our method, we choose the well-studied concept bottleneck models (CBMs) family as the baselines of our methods. 
As one of the main branches of concept-based models, CBMs first project the image feature to an intermediate concept bottleneck layer to obtain the concept activation vector. 
Each activation value indicates the presence of the corresponding concept in the input. 
The forward pipeline of vanilla CBM~\cite{cbm} can be formulated as:
\begin{align}
\label{eq: cbm}
\av =& f_C(f_I(\xv)),\\
\hat{y} =& f_{F}(\av)=W_F\cdot\av,
\end{align}
where $\av\in \mathbb{R}^M$ is the concept activation vector, $f_I: \mathbb{R}^{h\times w} \rightarrow \mathbb{R}^d$ is the image feature extractor maps the input $\xv$ into feature space, $f_C: \mathbb{R}^d \rightarrow \mathbb{R}^M$ is the concept projection layer that projects the image feature into concept activation vector, and $f_{F}: \mathbb{R}^M \rightarrow \mathbb{R}^{N_y}$ is the final linear classifier $W_F$ maps the concept activation into a final prediction.

\noindent \textbf{CLIP-CBM.} Vanilla CBM requires labor-intensive concept activation annotations to supervise the learning of concept activation. 
Contrastively, as illustrated in the blue part of Fig.~\ref{overview}, recent CLIP-based CBMs~\cite{labo, label-free, post-hoc-cbm} employ the pre-trained CLIP model to extract the image embeddings $E_I(x)$ and concept embeddings $E_T(c)$, respectively.
Subsequently, CLIP-CBM obtains the concept activations $\av$ by computing the similarity between image and concept embeddings, formulated as:
\begin{align}
\label{eq: clip-cbm}
a_i =& \text{sim}(E_I(x), E_T(c_i)), \\
\av =& \left[ a_1, a_2, ...,a_M \right], 
\end{align}
where $\text{sim}$ indicates the cosine similarity.

\begin{table*}[tp!]
\centering
\resizebox{0.9\textwidth}{!}{
    \begin{tabular}{l|c|c|cc|cc|cc|cc}
        \toprule[1pt]
        &&&\multicolumn{2}{c|}{CUB-Painting}  & \multicolumn{2}{c|}{AwA2-clipart}  & \multicolumn{2}{c|}{LADA-Sculpture} & \multicolumn{2}{c}{LADV-3D}\\
        Model &Concept &Method&ID &OOD  &ID &OOD &ID &OOD  &ID &OOD    \\\midrule
        \gg{CLIP ZS~\cite{clip}}&\gg{\multirow{2}{*}{\ding{55}}}  &\gg{\multirow{2}{*}{\ding{55}}} &\gg{62.21} &\gg{52.77}  &\gg{95.70} &\gg{90.26} &\gg{91.26} &\gg{82.05}&\gg{71.82}&\gg{66.29}  \\
        \gg{CLIP LP~\cite{clip}}& & &\gg{82.00} &\gg{61.40} &\gg{97.11} &\gg{86.75} &\gg{96.81}&\gg{74.40} &\gg{93.68}&\gg{63.81}\\\midrule
        \multirow{2}{*}{CLIP-CBM}&\multirow{2}{*}{human}  &baseline &78.51 &50.54 &95.69 &81.91 &96.66&70.44&92.21&60.64 \\
        &  &\cellcolor{gray!20}\textbf{+DDO} &\cellcolor{gray!20}\textbf{78.70}&\cellcolor{gray!20}\textbf{55.53}&\cellcolor{gray!20}\textbf{95.71} &\cellcolor{gray!20}\textbf{83.72} &\cellcolor{gray!20}\textbf{96.77}&\cellcolor{gray!20}\textbf{75.76}&\cellcolor{gray!20}\textbf{92.59}&\cellcolor{gray!20}\textbf{63.51}\\\midrule
        \multirow{2}{*}{PCBM$\dag$~\cite{post-hoc-cbm}}&\multirow{2}{*}{ConceptNet}  &baseline & 75.85&54.41&97.17&84.77&97.60&76.69&94.71&65.88   \\
        &  &\cellcolor{gray!20}\textbf{+DDO} &\cellcolor{gray!20}\textbf{76.48}&\cellcolor{gray!20}\textbf{57.50}&\cellcolor{gray!20}\textbf{97.19}&\cellcolor{gray!20}\textbf{86.58}&\cellcolor{gray!20}\textbf{97.64}&\cellcolor{gray!20}\textbf{79.74}&\cellcolor{gray!20}\textbf{94.82}&\cellcolor{gray!20}\textbf{68.33} \\\midrule
        \multirow{2}{*}{LaBO$\ddagger$~\cite{labo}}&\multirow{2}{*}{LLM}   &baseline &81.91 &56.24&97.14 &84.15&97.41 &74.56&99.90 &63.17\\
        &  &\cellcolor{gray!20}\textbf{+DDO} &\cellcolor{gray!20}\textbf{82.34}&\cellcolor{gray!20}\textbf{59.60} &\cellcolor{gray!20}\textbf{97.26} &\cellcolor{gray!20}\textbf{87.66}&\cellcolor{gray!20}\textbf{98.12} &\cellcolor{gray!20}\textbf{80.00} &\cellcolor{gray!20}\textbf{99.93} &\cellcolor{gray!20}\textbf{68.01} \\
        \bottomrule[1pt]
    \end{tabular}}
    \vspace{-3mm}
\caption{Performance on four single unseen domain benchmarks. For comparison, we list the performance of prevailing CBMs as baselines and report the results of integrating our proposed DDO regularizer into these baselines. $\dag$ indicates re-implemented with CLIP backbone. $\ddagger$ means visual concepts are obtained with our re-implemented LLM prompts. ID is the performance on photo domains, and OOD is the generalization performance on other domains.}
\label{Table:singleunseen}
\end{table*}

The objective of CLIP-CBM is:
\begin{align}
\label{eq: ce}
\mathcal{L}_\text{CE}(\hat{y},y) =  {\mathbb{E}}_{(\boldsymbol{x},y) \sim \mathcal{D}_{train}}\left[\mathcal{L}\left(W_F\cdot \av, y\right)\right],
\end{align}
where $\mathcal{L}_\text{CE}(\hat{y},y)$ is the cross-entropy loss.

\subsection{Domain Descriptors Orthogonality Loss}
\label{ddoloss}
As mentioned in Sec.~\ref{introduction}, CLIP-CBMs suffer from the OOD generalization problem due to the negative impact of domain-specific concepts. 
To erase the negative influence of domain-specific concepts, we encourage the learnable class-concept correlation matrix ($\ie$ the weight of the final linear classification layer), to be orthogonal to the language-guided synthesized domain-specific concept activations.

Following previous works~\cite{csd,li2018deep}, we decompose the concept activation $\av$ into two components, the \textit{domain-specific} concept activation vector $\av_{sp}$ and the \textit{domain-shared} concept vector $\av_{sh}$, represented as: 
\begin{align}
\label{eq: ce}
\av = \av_{sp} + \av_{sh},
\end{align}

Our goal is to erase the impact of domain-specific concepts to the final predictions, formulated as:
\begin{align}
\label{eq: 2}
 W_F\cdot \av_{sp}= 0.
\end{align}

Based on the observations in Sec.~\ref{empirical}, we can simulate the $\av_{sp}$ via computing the similarity between language-guided domain shifts and the concepts.
Specifically, the language-guided class-level domain shifts $\Delta t_{py}$ are the subtraction between text embeddings of prompts of possible unseen domains and the training domain ($\eg$ ``A sketch of an apple'' - ``A photo of an apple'').
\begin{align}
\Delta t(p_i,y) =& E_T([p_i,y]) - E_T([p_{train},y]), \label{eq: domain shift}\\
&\widehat{\av_{\text{sp}}(p_i, y)} = E_T(c) \cdot \Delta t(p_i,y)  \label{eq: domain-specific-activation}
\end{align}
where $\widehat{\av_{\text{sp}}(p_i, y)}$ is simulated domain-specific concept activation and $p_{i}, p_{train}, y$ are the unseen domain descriptor, training domain descriptor and class, respectively.

For all possible unseen domains $\mathcal{P}$ and candidate classes $\mathcal{Y}$, based on Eq.~\eqref{eq: 2} and Eq.~\eqref{eq: domain-specific-activation}, we derive our proposed domain descriptor orthogonality (DDO) loss:
\begin{align}
\label{eq: ce}
\mathcal{L}_{\text{DDO}} = \mathbb{E}_{(p_i, y) \sim \mathcal{P} \times \mathcal{Y}} \left[ \left| W_F \cdot \widehat{\av_{\text{sp}}(p_i, y)} \right| \right]
\end{align}
The forward process of DDO is illustrated in the yellow part of Fig.~\ref{overview}. Note that the input to the DDO loss is independent of the current samples and only acts as a regularization term applied to the $W_F$. Therefore, it can be plugged into various CBMs.
The final loss is the combination of classification loss and DDO loss, stated as:
\begin{align}
\label{eq: final}
\mathcal{L} = \mathcal{L}_{\text{CE}} + \lambda \mathcal{L}_{\text{DDO}}
\end{align}

\subsection{Generate Domain Descriptors}
\label{Generate domain descriptors}
To simulate various unseen visual domains as much as possible, we need to write numerous domain descriptors.
However, hand-writing these domain descriptors can be costly, and does not scale to large numbers of unseen domains. 
We can automatically construct this domain descriptor set $\mathcal{P}=\{p_i\}_{i=1}^{N_p}$ by prompting a large language model, such as GPT-3.5~\cite{chatgpt2023}, to list possible unseen visual domains excepts for the training domain.
We prompt the large language model with the input:
\begin{quote}
\textit{Q: Please list visual domains in short phrases as much as possible.} \\
\textit{A: Here are some visual domains in short phrases: real-world photography, clipart illustrations, 3D renders...}
\end{quote}
The generated $N_p$ domain descriptors comprise the domain descriptor set $\mathcal{P}$. The full generation list and further implementation details can be found in the Appendix \ref{appendix:implementation}.

\section{Experiments}

\subsection{Datasets and Baselines}
We conduct experiments on seven domain adaptation benchmarks. 
Notably, CUB-Painting, PACS, OfficeHome and DomainNet are the previous common benchmarks, while AwA2-clipart, LADA-Sculpture, and LADV-3D are newly proposed in the paper. 

\noindent\textbf{Single unseen domain benchmarks.} There are some classical image classification datasets with attribute annotations, which are widely applied for the evaluation of concept-based models. For instance, \textit{CUB-Painting} is a fine-grained bird classification dataset that consists of two visual domains, the photo domain~\cite{cub} and the painting domain~\cite{cub-painting}. Each image has an attribute annotation vector. 
Following CUB-Painting, we collect images with various other styles for photographic images from Animals with Attributes 2 (AwA2)~\cite{awa2}, LAD-animal~\cite{lad} and LAD-vehicle~\cite{lad}. 
Specifically, we introduce three new benchmarks, namely \textit{AwA2-clipart}, \textit{LADA-Sculpture} and \textit{LADV-3D}, which focused on photo$\rightarrow$clipart, real$\rightarrow$sculpture and real$\rightarrow$3D model, respectively.



\noindent\textbf{Multiple unseen domain benchmarks.} There are also some classical domain adaptation benchmarks with multiple visual domains. 
\textit{PACS}~\cite{pacs} is a domain adaptation benchmark with 7 classes and 4 visual styles, including art, cartoon, photo, and sketch. 
\textit{OfficeHome}~\cite{officehome} contains 4 domains, $\ie$ art, clipart, product, realworld, where each domain consists of 65 categories.
\textit{DomainNet}~\cite{domainNet} is a domain adaptation dataset containing 345 common classes from six visual domains, $\ie$ real, clipart, infograph, painting, quickdraw, and sketch. 
Following previous single domain generalization approaches~\cite{single1,single2}, we train on one source domain and evaluate on the other target domains. For each source domain, we report the average generalization accuracy on all target domains.

\begin{figure*}[!tb]
	\centering 
	\includegraphics[width=1.0\textwidth]{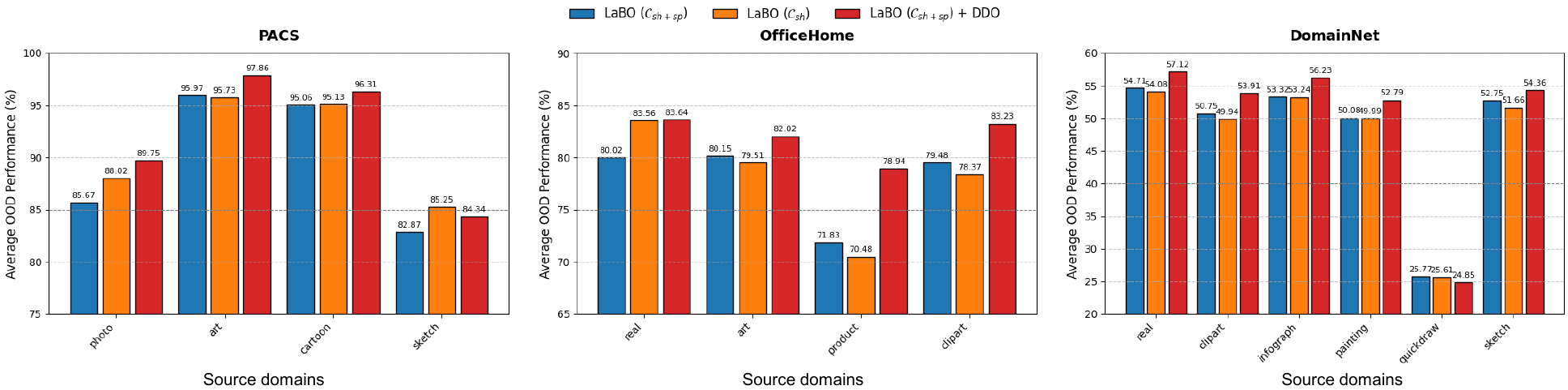}
        \vspace{-5mm}
	\caption{OOD performance on three multiple unseen domain generalization benchmarks, PACS, OfficeHome and DomainNet.}\label{radar}
\end{figure*}

\noindent\textbf{Baselines.} We compare our proposed method with the following baseline approaches. 
1) Black-box classifier: 
\textit{CLIP-ZS}~\cite{clip} uses cosine similarity between class textual representation and image representation in CLIP shared space.
\textit{CLIP-LP}~\cite{clip} fits a linear classifier on top of the frozen CLIP image encoder.
2) CLIP-based CBMs:
\textit{CLIP-CBM} leverages human-written attributes to serve as a concept bank and utilizes CLIP to compute concept activation label.
\textit{PCBM}~\cite{post-hoc-cbm} uses ConceptNet~\cite{conceptNet} to retrieve concepts relevant to these classes and automatically construct the concept bank. Original PCBM employs different visual backbones on different datasets. For a fair comparison, we re-implement the PCBM with the CLIP backbone.
\textit{LaBO}~\cite{labo} applies the LLM to generate abundant visual concepts in contrast to PCBM, and we employ a two-layer classifier.
For each CBM baseline, we keep their original objectives to solve the optimization.

\subsection{Implementation Details}
We prompt GPT3.5-turbo to generate 200 domain descriptors. These domain descriptors are used across all datasets and $\lambda$ is set as 1. We pre-process the domain descriptors into text embeddings by CLIP text encoder. The DDO loss is added at the final classification layer of the CBMs in a plug-in manner.
Following LaBO~\cite{labo}, we utilize the official CLIP ViT-L/14\footnote{\url{https://github.com/openai/CLIP}} by OpenAI as the default vision backbone to extract image embeddings.
We train the linear function $W_F$ using the Adam~\cite{kingma2014adam} optimizer. All experiments are conducted on a single RTX3090 GPU. More implementation details and experiments with other vision backbones are listed in the Appendix \ref{appendix:implementation} \& \ref{appendix:ablation}.
\begin{figure*}[!tb]
	\centering 
	\includegraphics[width=1.0\textwidth]{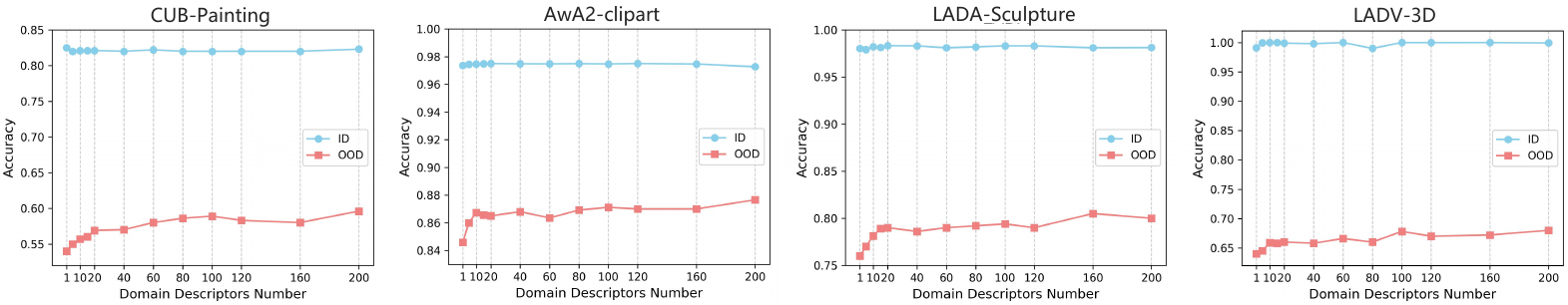}
        \vspace{-5mm}
	\caption{Ablation studies for the impact of the number of domain descriptors. For each quantity, we randomly selected domain descriptors from a total of 200 domain descriptors and averaged the results over five random selections. Results of DomainNet are shown in the Appendix \ref{appendix:ablation}.}\label{number}

\end{figure*}

\subsection{Generalization to single unseen domain}
We first evaluate our model on single unseen domain benchmarks which focused on photo $\rightarrow$ others generalization. Table.~\ref{Table:singleunseen} shows in-distribution (ID) and out-of-distribution (OOD) accuracy on CUB-Painting, AwA2-clipart, LADA-Sculpture and LADV-3D.
For all benchmarks, DDO loss is able to preserve or slightly improve the ID accuracy of each CBM while significantly improving their OOD accuracy (average 3 points improvements). We attribute the slight improvements in ID performance to the minor distribution differences between the training and test sets, even in the same visual domain. 
Moreover, LaBo with DDO loss can achieve competitive and superior performance on both ID and OOD performance with black-box CLIP-LP and narrow the gap with CLIP-ZS on OOD accuracy. 
These results demonstrate that our proposed LanCE design can effectively improve the generalization capabilities without sacrificing interpretability or classification accuracy.
Besides, we notice that all models' generalization performance to sculpture and 3D model images on the LADA-Sculpture and LADV-3D benchmark is limited, indicating generalization from 2D to 3D remains challenging. 

\subsection{Generalization to multiple unseen domains}
To assess the ability to generalize to multiple unseen domains and the generalization capability when different domains act as the source domain, we conduct experiments on PACS, OfficeHome, and DomainNet.
\begin{table}[tp!]
\centering
\resizebox{0.45\textwidth}{!}{
    \begin{tabular}{l|c|c|c|c|c}
        \toprule[1pt]
        Model&Method&CUB-P  & AwA2-c  &LADA-S &LADV-3D\\\midrule
        \multirow{3}{*}{LanCE}   &baseline  &56.24 &84.15 &74.56 &63.17\\
        &\textbf{+DDO(IR)} &57.60&85.50&77.70&65.46 \\
        &\textbf{+DDO} &\textbf{59.60}  &\textbf{87.66} &\textbf{80.00} &\textbf{68.01} \\
        \bottomrule[1pt]
    \end{tabular}}
    \vspace{-3mm}
\caption{Ablation studies for the effect of relevance of the domain descriptors. +DDO(IR) only use the domain-irrlevant descriptors while +DDO use all domain descriptors.}
\label{Table:abforremoveing}
\end{table}

\begin{figure*}[!tb]
	\centering 
	\includegraphics[width=1.0\textwidth]{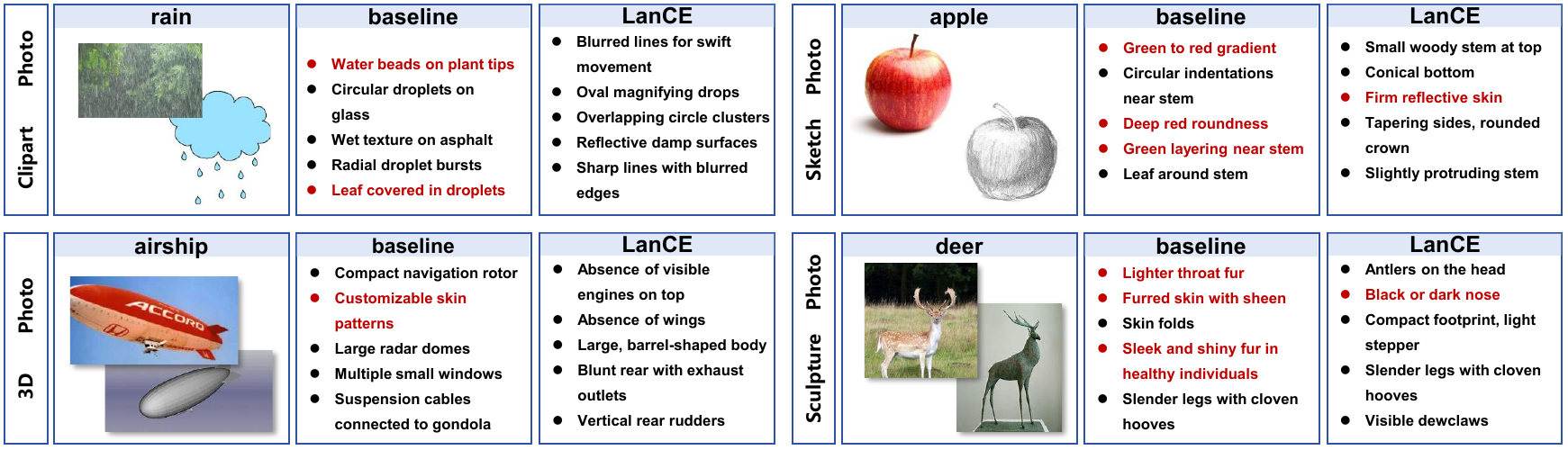}
        \vspace{-7mm}
	\caption{Qualitative examples about the top-5 concepts, ranking by their weights in the final linear layer $W_F$. Baseline indicates the results of the original LaBo. Domain-specific concepts are hightlighted in \textcolor[RGB]{192,0,0}{red}. More results are listed in Appendix \ref{appendix:qualitative}.}\label{qualitative}
\end{figure*}

\noindent\textbf{Comparison with only using domain-shared concepts.} In the absence of a pre-defined human-written concept set, we employ LLM to generate a concept set $\mathcal{C}_{sp+in}$ for each dataset following LaBO~\cite{labo}.
This set contains both domain-specific concepts and domain-shared concepts. The main motivation of this paper is to mitigate the negative impact of domain-specific concepts.  A natural question that arises is why not simply query the LLM to generate only domain-shared concepts.
 To explore the differences between this domain-shared approach and our DDO loss, we introduce an additional baseline, where the LLM is queried exclusively to generate a domain-shared concept set $\mathcal{C}_{sh}$.
 Specifically, we adopt the single-domain generalization evaluation setting where one source domain is used for training, and the OOD performance is averaged across other domains.  As shown in Fig.~\ref{radar}, only using $\mathcal{C}_{sh}$ performs well only in certain cases, such as when the source domains in the PACS dataset are photo or sketch. We attribute this to the fact that the manifestation of domain-shared concepts varies across different domains. For instance, deer antlers in photo (realistic) and cartoon (simplified or exaggerated) domains differ significantly. Moreover, we empirically observe that prompt-based methods are highly sensitive to the wording of human-written prompts. Different prompts lead to significant performance variations, yet they share a common limitation: none consistently improve performance across diverse source domains. In contrast, DDO is more flexible. It adaptively erases the influence of domain-specific concepts while suppressing domain-specific information in domain-shared concept activations, leading to stable improvements across a variety of source domains.

\subsection{Ablation Studies}
To explore the impact of domain descriptors on final performance, we conduct comprehensive ablation studies, with LaBO as the baseline. Specifically, we compare results with different numbers of domain descriptors and investigate the role of distinct domain descriptors. 

\noindent\textbf{The impact of the numbers of domain descriptors.}
We evaluate our method with regard to the number of domain descriptors $N_p$ as shown in Fig.~\ref{number}. 
The ID accuracy remains almost constant as the number of domain descriptors increases, while the OOD accuracy gradually improves with an increasing number of domain descriptors, though with diminishing returns. Specifically, when the number of domain descriptors is below 20, each additional descriptor noticeably enhances OOD accuracy. However, as the number of domain descriptors continues to increase beyond this point, the improvement in OOD accuracy slows progressively. The results shows that 100 domain descriptors are enough to achieve decent performance.


\noindent \textbf{The impact of the relevance of domain descriptors.}
To investigate which parts of 200 domain descriptors contribute the most to OOD accuracy improvements, we manually split all domain descriptors into two components, domain-relevant descriptors and domain-irrelevant descriptors based on their relevance with the test unseen domains.
For example, for the LADA-Sculpture dataset, domain-relevant domain descriptors are ``sculpture'', ``statue'', and ``furniture'', referring to the generated word cloud mentioned in Fig.~\ref{empiricalObservation} and domain-irrelevant parts are the remaining ones.
Detailed splits for each dataset are listed in the Appendix \ref{appendix:ablation}. 
Results are shown in the Table.~\ref{Table:abforremoveing}.
DDO(IR) indicates that we only use the domain-irrelevant domain descriptors to compute the DDO loss.
This evaluation shows that domain-irrelevant domain descriptors can also slightly prompt the OOD accuracy on the unseen domains, although relatively fewer than the domain-relevant counterparts.
It indicates those domain descriptors are not fully orthonormal, also demonstrating that the DDO loss has the potential to generalize to pure unseen visual domains that are not mentioned in the domain descriptors set $\mathcal{P}$.

\subsection{Qualitative Results}
To evaluate the effectiveness of our proposed LanCE method, we further visualize the Top-5 concepts highly related to randomly selected classes within different unseen domains as illustrated in Fig.~\ref{qualitative}.
The qualitative results demonstrate that our proposed method can significantly decrease the association between the final prediction and the domain-specific concepts, in other words, erasing the impact of domain-specific concepts on the final output. More qualitative results are shown in the Appendix \ref{appendix:qualitative}.

\section{Conclusion}
In this paper, we propose a language-guided concept-erasing (LanCE) framework that can effectively mitigate the biased association between domain-specific concepts and the final output.
We first empirically demonstrate that the pre-trained CLIP can interpret the visual domain shifts into language. And these language-guided domain shifts can distinguish the domain-specific concepts via similarity within the CLIP embedding space. Based on these observations, we introduce a plug-in domain descriptor orthogonality (DDO) loss to erase the negative impact of domain-specific concepts, and that can significantly improve the generalization capabilities of prevailing concept bottleneck models (CBMs). We prompt the large language models (LLMs) to generate a bunch of domain descriptors to simulate numerous possible unseen domains.
Moreover, considering that the current domain generalization community focuses more on generalizing to artistic unseen domains, we collect three new benchmarks that includes more difficult scenarios, $\ie$ 2D$\rightarrow$3D, real animals$\rightarrow$sculptures. Extensive experiments demonstrate the efficacy of our LanCE.

\section{Acknowledgement}
This work was supported in part by the National Natural Science Foundation of China under Grant U21B2006; in part by Shaanxi Youth Innovation Team Project; in part by the Fundamental Research Funds for the Central Universities QTZX24003 and QTZX22160; in part by the 111 Project under Grant B18039;
Hao Zhang acknowledges the support of NSFC (62301384); Excellent Young Scientists Fund (Overseas); Foundation of National Key Laboratory of Radar Signal Processing under Grant JKW202308. Zhengjue Wang acknowledges the support of NSFC (62301407).

{
    \small
    \bibliographystyle{ieeenat_fullname}
    \bibliography{main}
}
\appendix
\onecolumn
\section{New Datasets}
\label{appendix:datasets}
\begin{figure*}[!thb]
	\centering 
	\includegraphics[width=1.0\textwidth]{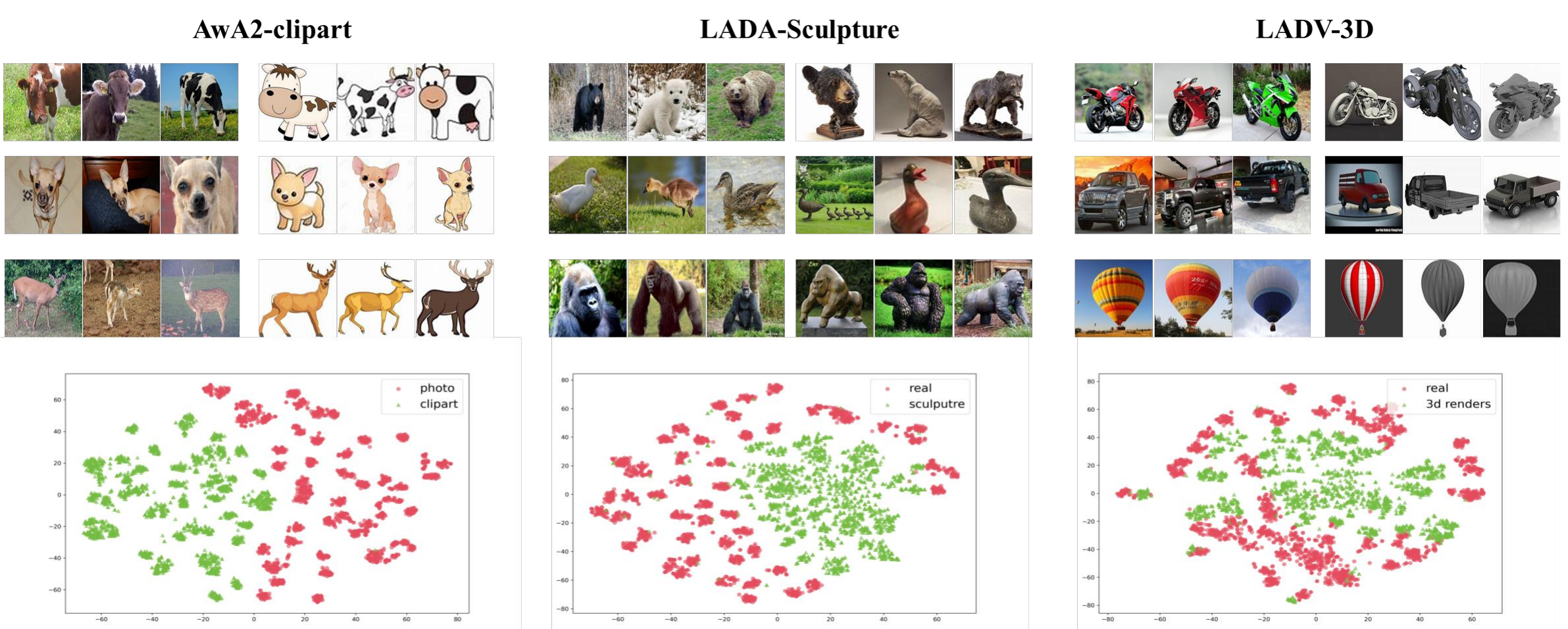}
        \vspace{-6mm}
	\caption{Some randomly selected samples and CLIP feature TSNE visualization of all samples on our proposed three benchmarks.}\label{dataset}
\end{figure*}

\textbf{Dataset Statistics.} We collect three new domain adaptation datasets, $\ie$ AwA2-clipart, LADA-Sculpture, and LADV-3D based on previous attribute-annotated datasets~\cite{awa2,lads} in this paper. Table.~\ref{Table:Dataset} are some basic dataset statistics.  
Fig.~\ref{dataset} has shown some samples from each dataset.
To visualize the domain shifts in these three new datasets, we use CLIP ViT-L/14 to extract image features for images in these datasets and use the TSNE tools to visualize all image features.
\begin{table*}[thb!]
\centering
\resizebox{0.65\textwidth}{!}{
    \begin{tabular}{l|l|l|l}
        \toprule[1pt]
        Dataset Statistics& AwA2-clipart  & LADA-Sculpture & LADV-3D \\\midrule
        Data field  & animals &animals &vehicles  \\
        Visual domains  & photo,clipart & real,sculputre & real, 3d renders  \\
        Number of images &37328,5319&13240,2162&17080,3587 \\
        Number of categories &50&50&50 \\
        \bottomrule[1pt]
    \end{tabular}}

\caption{Some dataset statistics of AwA2-clipart, LADA-Sculpture, LADV-3D.}
\label{Table:Dataset}
\end{table*}

\section{Empirical studies}
\label{appendix:empirical}
\begin{figure*}[!thb]
	\centering 
	\includegraphics[width=1.0\textwidth]{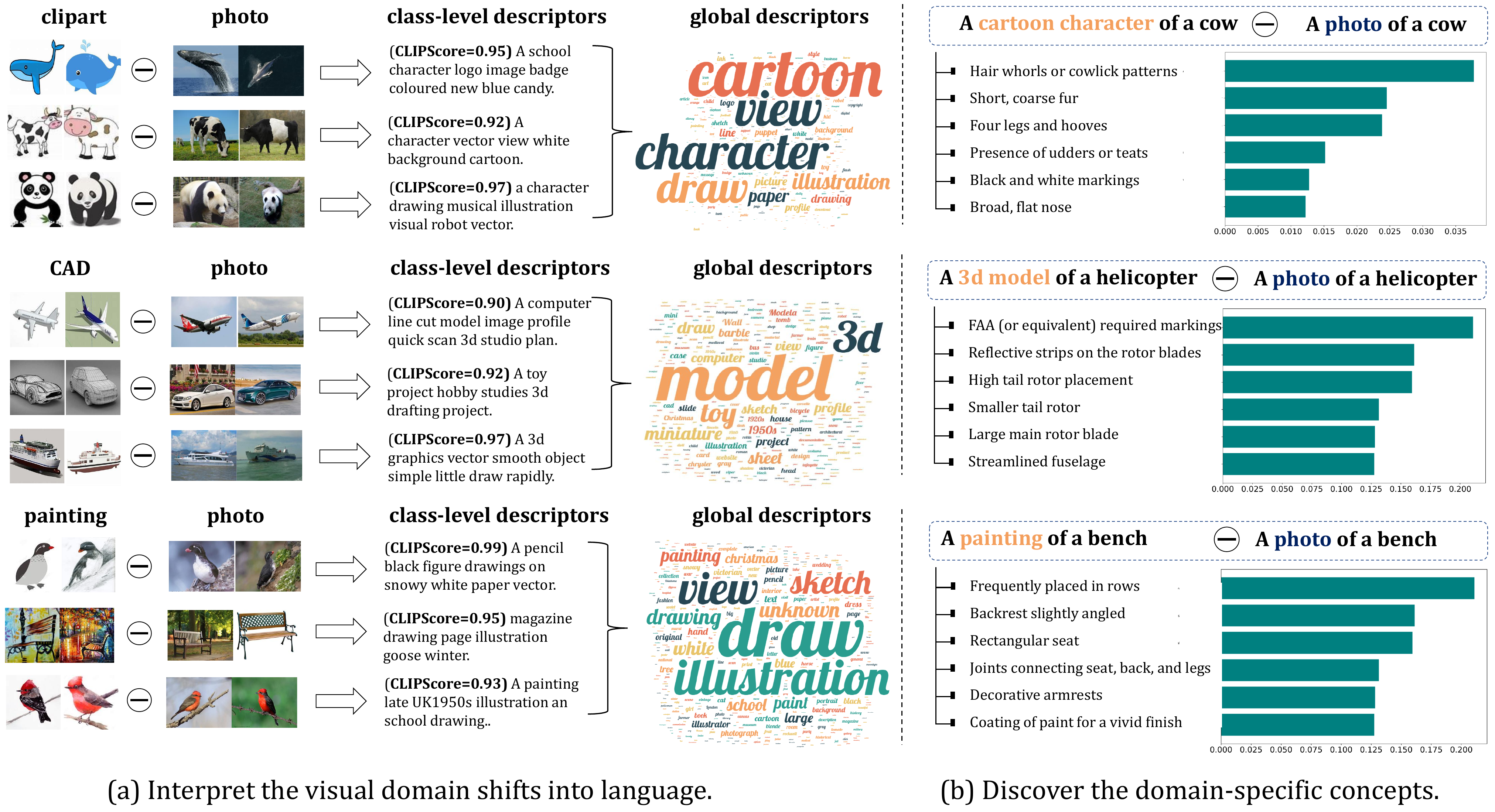}
        \vspace{-6mm}
	\caption{Empirical studies on more domain shifts.}\label{empiricalObservation1}
\end{figure*}

This paper's key insights are derived from our empirical studies that 
the VLMs can interpret the visual domain shifts into language, in other words, visual domain shifts and descriptions of different domains are consistent in the VLM embedding space. 
The main paper lists the results between two unseen visual domains ($\ie$ ``sketch'' and ``sculpture'') and the training domain (``photo''). In this section, we list more results between unseen visual domains like ``clipart'', ``3d model'' and ``painting'' in Fig.~\ref{empiricalObservation1}.
Similar to the main paper, the class-level descriptors generated by ConZIC~\cite{zeng2023conzic} describe the visual difference. The style descriptions for the common photo domain (subtrahend) are generally implicit in the training caption corpus, thus, the class-level descriptors prominently consist of style descriptions about the unseen visual domains (minuend).  
To better visualize the global semantic direction, we aggregate all class-level descriptors into a final word cloud format, where prominent words represent the main visual direction between the two visual domains. 
For instance, semantical words ``cartoon, view, character, draw, illustration'' indicate the main style direction of the  ``clipart'' domain. 
This demonstrates that the visual domain shifts can be approximated by some domain-related style descriptors, while these style descriptors can used to compute class-level textual domain differences.`

Then, we can utilize these textual domain differences to discover the domain-specific concepts.
Concretely, we can concatenate these domain descriptors with specific classes in a prompt style, such as ``A cartoon character of a cow'', denotes as  $t_{tgt}$. 
Similarly, we can construct a prompt for photo domains like ``A photo of a cow'', denoted as $t_{src}$. For each discriminative visual concept $c_i$, we can compute the similarity $s_i$ with this class-level textual domain shifts in the CLIP embedding space, as:
\begin{align}
\Delta t =& E_T(t_{tgt})-E_T(t_{src}) \label{domain shift embeddings} \\
s_i =& E_T(c_i)\cdot \Delta t , \label{domain-specific concept activation}
\end{align}
where $E_T$ means CLIP text encoder. Finally, we can get all concept activation $\{s_i\}_{i=1}^M$.
We empirically find that those concepts with higher similarity are often domain-specific concepts that exhibit substantial variation between two domains. Therefore, $S=[s_1,...,s_M]$ can viewed as a domain-specific concept activation $a_{sp}$. These observations provide the key insights of this paper.



        

\begin{figure*}[!tb]
	\centering 
	\includegraphics[width=1.0\textwidth]{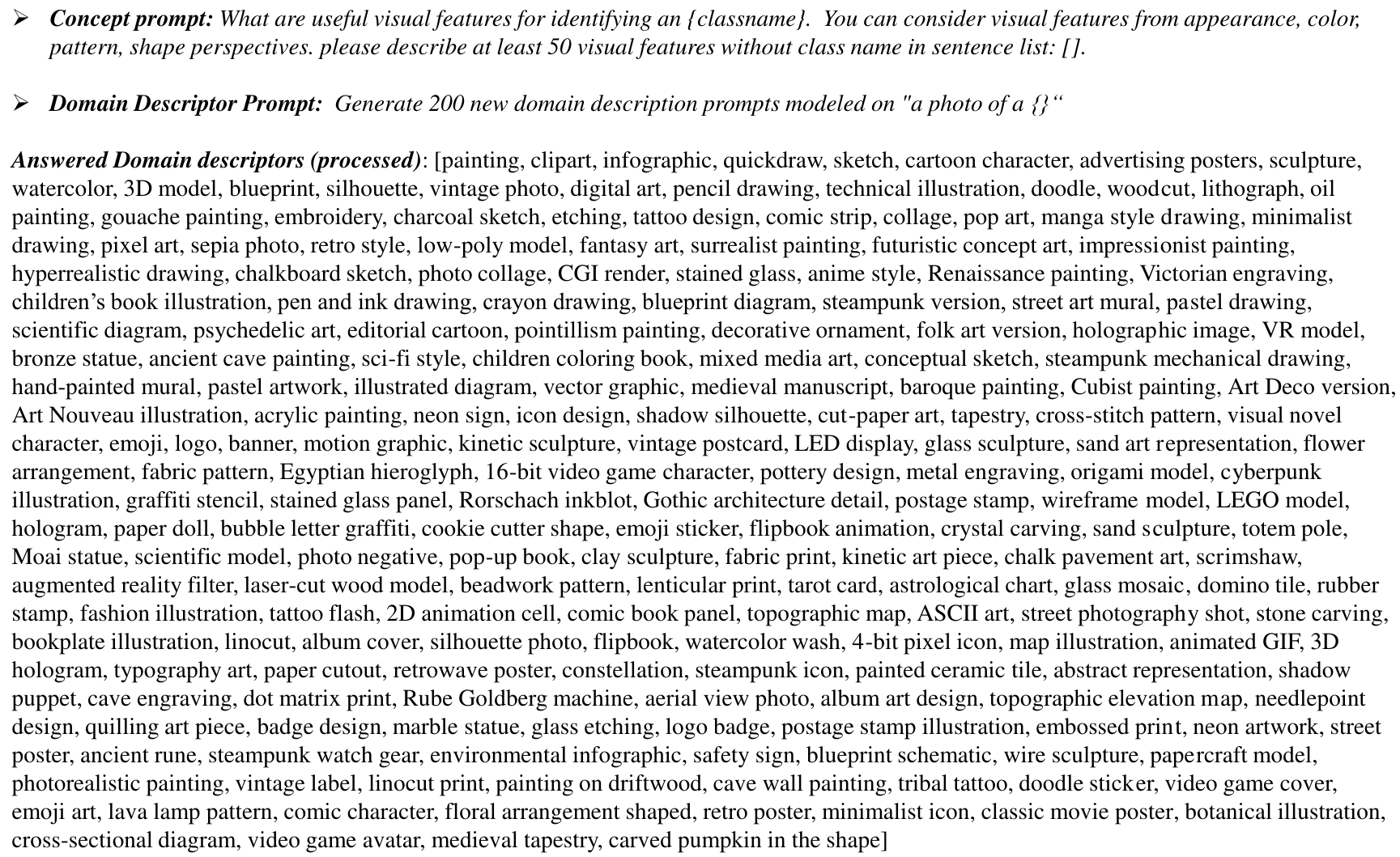}
	\caption{LLM prompts to generate visual concepts and domain descriptors and detailed generated domain descriptor list.}\label{prompt}
\end{figure*}


\section{More implementation details of DDO loss}
\label{appendix:implementation}
In this section, we will introduce more implementation details about our proposed LanCE framework.
\subsection{Generating domain descriptors} 
Based on the empirical observations in Sec.~\ref{appendix:empirical}, we aim to achieve generalization across a wide range of unseen visual domains by leveraging a large language model (LLM) to generate domain descriptors $\mathcal{P}$. 
A detailed list is shown in Fig.~\ref{prompt}.

\subsection{Textual domain shifts embeddings}
After we get all domain descriptors $\mathcal{P}=\{p_i\}_{i=1}^{N_{p}}$, we can use these domain descriptors to compute textual domain shift embeddings based on Eq.~\eqref{domain shift embeddings}. 
Specifically, we set the training domain as ``photo'' and compute the difference between two domain-related class prompts. Finally, we can get all class-level textual domain shifts $[\Delta t(p_i,y)]_{N_p \times N_y}$. 
\subsection{Details of DDO loss}
DDO aims to encourage the orthogonality between all domain-specific concept activations $[\boldsymbol{a}_{sp}(p_i,y)]_{N_p \times N_y}$ and class prototype concept activations ($\ie$ linear weight $W_F$). The computation of $\boldsymbol{a}_{sp}(p_i,y)$ as:
\begin{align}
\boldsymbol{a}_{sp}(p_i,y) = [\Delta t(p_i,y) \cdot E_T(c_1),...,\Delta t(p_i,y) \cdot E_T(c_M)]
\end{align}
Notably, the DDO loss is independent of specific input samples, as all $[\boldsymbol{a}_{sp}(p_i,y)]{N_p \times N_y}$ are processed by $W_F$ collectively with each batch of image samples.


\section{More ablation studies}
\label{appendix:ablation}
\textbf{Effect of CLIP backbone.} Table.~\ref{Table:singleunseenAppendix} and Table.~\ref{Table:multipleunseendomainAppendix} have shown the detailed results with CLIP ViT-B/32 and CLIP ViT-L/14, demonstrating that our proposed DDO regularizer can improve the OOD accuracy across different CLIP image backbones.

\begin{figure*}[!thb]
	\centering 
	\includegraphics[width=0.75\textwidth]{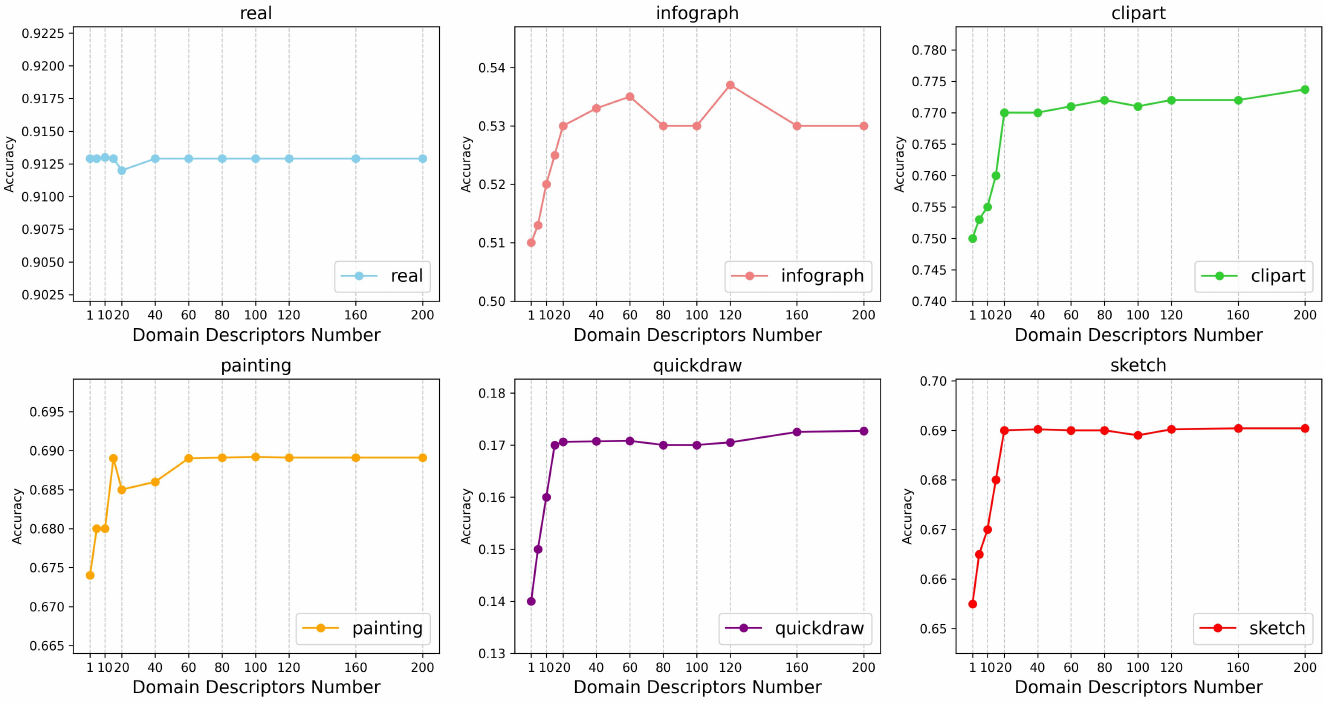}
	\caption{Ablation studies about the numbers of the domain descriptors on DomainNet.}\label{domainNet_ablation}
\end{figure*}

\noindent\textbf{Effect of the numbers of domain descriptors.}
Fig.~\ref{domainNet_ablation} provides the ablation studies on DomainNet. Similar to other results on other datasets, the OOD accuracy gradually improves with an increasing number of domain descriptors.

\noindent\textbf{Effect of relevance of domain descriptors.}
Table.~\ref{Table:relevant} has shown some relevant keywords about each benchmark. 
We remove domain descriptors containing these keywords and investigate the contribution of remaining domain-irrelevant domain descriptors to improving OOD accuracy.
Fig.~\ref{Table:abforremoveingDomainNet} shows the results on DomainNet.

\noindent\textbf{Computation complexity.} To evaluate the gained computation complexity brought by DDO loss, we list the comparison results of FLOPs, trainable parameters, and estimated memory usage.
Results are shown in Table.~\ref{Table:computation}, as we can see, the gained computation complexity is minor and almost negligible.
It demonstrates that our proposed DDO loss is a plug-in loss that can be applied to many concept-based models without changing model architecture and increasing too much computation cost.

\begin{table}[tp!]
\centering
\resizebox{0.55\textwidth}{!}{
    \begin{tabular}{l|c|c|c}
        \toprule[1pt]
        Model&FLOPs (in billions)&Parameters  &Memory Usage (in MB) \\\midrule
        baseline  &13285.1 &63224 &918.0\\
        \textbf{LanCE} &13287.7 &63224 &1038.1 \\
        \bottomrule[1pt]
    \end{tabular}}
    \vspace{-3mm}
\caption{Comparison of computation complexity, including FLOPs, trainable parameters, and estimated memory usage.}
\label{Table:computation}
    \vspace{-3mm}
\end{table}

\begin{table*}[tp!]
\centering
\resizebox{0.9\textwidth}{!}{
    \begin{tabular}{l|c|c|cc|cc|cc|cc}
        \toprule[1pt]
        
        &&&\multicolumn{2}{c|}{CUB-Painting}  & \multicolumn{2}{c|}{AwA2-clipart}  & \multicolumn{2}{c|}{LADA-Sculpture} & \multicolumn{2}{c}{LADV-3D}\\
        Model &Concept &Method&ID &OOD  &ID &OOD &ID &OOD  &ID &OOD    \\\midrule
        \multicolumn{11}{c}{CLIP ViT-B/32} \\\midrule
        \gg{CLIP ZS~\cite{clip}}&\gg{\multirow{2}{*}{\ding{55}}}  &\gg{\multirow{2}{*}{\ding{55}}} &\gg{65.27} &\gg{40.14}  &\gg{91.50} &\gg{67.44} &\gg{87.17} &\gg{48.98}&\gg{84.36}&\gg{50.07}  \\
        \gg{CLIP LP~\cite{clip}}& & &\gg{51.59} &\gg{44.57} &\gg{91.72} &\gg{79.32} &\gg{89.46}&\gg{65.12} &\gg{68.28}&\gg{60.49}\\\midrule
        \multirow{2}{*}{CLIP-CBM}&\multirow{2}{*}{human}  &baseline &61.05 &35.08 &92.38 &62.14 &94.52&48.89&87.85&53.25 \\
        &  &\cellcolor{gray!20}\textbf{+DDO} &\cellcolor{gray!20}\textbf{62.88}&\cellcolor{gray!20}\textbf{39.51}&\cellcolor{gray!20}\textbf{90.95} &\cellcolor{gray!20}\textbf{66.29} &\cellcolor{gray!20}\textbf{95.05}&\cellcolor{gray!20}\textbf{53.70}&\cellcolor{gray!20}\textbf{87.29}&\cellcolor{gray!20}\textbf{55.42}\\\midrule
        \multirow{2}{*}{PCBM$\dag$~\cite{post-hoc-cbm}}&\multirow{2}{*}{ConceptNet}  &baseline & 58.31&38.83&93.41 &\textbf{71.78} &95.31&55.64&90.11&56.87   \\
        &  &\cellcolor{gray!20}\textbf{+DDO} &\cellcolor{gray!20}\textbf{58.81}&\cellcolor{gray!20}\textbf{39.91}&\cellcolor{gray!20}\textbf{92.58}&\cellcolor{gray!20}69.11&\cellcolor{gray!20}\textbf{95.76}&\cellcolor{gray!20}\textbf{57.96}&\cellcolor{gray!20}\textbf{90.03}&\cellcolor{gray!20}\textbf{57.18} \\\midrule
        \multirow{2}{*}{LaBO~\cite{labo}}&\multirow{2}{*}{LLM}   &baseline &67.57 &35.35&93.90 &62.60&96.70 &77.63&99.44 &56.00\\
        &  &\cellcolor{gray!20}\textbf{+DDO} &\cellcolor{gray!20}\textbf{67.83}&\cellcolor{gray!20}\textbf{37.28} &\cellcolor{gray!20}\textbf{94.50} &\cellcolor{gray!20}\textbf{71.70}&\cellcolor{gray!20}\textbf{98.27} &\cellcolor{gray!20}\textbf{79.69} &\cellcolor{gray!20}\textbf{99.13} &\cellcolor{gray!20}\textbf{58.88} \\\midrule
                \multicolumn{11}{c}{CLIP ViT-L/14}\\\midrule
                \gg{CLIP ZS~\cite{clip}}&\gg{\multirow{2}{*}{\ding{55}}}  &\gg{\multirow{2}{*}{\ding{55}}} &\gg{62.21} &\gg{52.77}  &\gg{95.70} &\gg{90.26} &\gg{91.26} &\gg{82.05}&\gg{71.82}&\gg{66.29}  \\
        \gg{CLIP LP~\cite{clip}}& & &\gg{82.00} &\gg{61.40} &\gg{97.11} &\gg{86.75} &\gg{96.81}&\gg{74.40} &\gg{93.68}&\gg{63.81}\\\midrule
        \multirow{2}{*}{CLIP-CBM}&\multirow{2}{*}{human}  &baseline &78.51 &50.54 &95.69 &81.91 &96.66&70.44&92.21&60.64 \\
        &  &\cellcolor{gray!20}\textbf{+DDO} &\cellcolor{gray!20}\textbf{78.70}&\cellcolor{gray!20}\textbf{55.53}&\cellcolor{gray!20}\textbf{95.71} &\cellcolor{gray!20}\textbf{83.72} &\cellcolor{gray!20}\textbf{96.77}&\cellcolor{gray!20}\textbf{75.76}&\cellcolor{gray!20}\textbf{92.59}&\cellcolor{gray!20}\textbf{63.51}\\\midrule
        \multirow{2}{*}{PCBM$\dag$~\cite{post-hoc-cbm}}&\multirow{2}{*}{ConceptNet}  &baseline & 75.85&54.41&97.17&84.77&97.60&76.69&94.71&65.88   \\
        &  &\cellcolor{gray!20}\textbf{+DDO} &\cellcolor{gray!20}\textbf{76.48}&\cellcolor{gray!20}\textbf{57.50}&\cellcolor{gray!20}\textbf{97.19}&\cellcolor{gray!20}\textbf{86.58}&\cellcolor{gray!20}\textbf{97.64}&\cellcolor{gray!20}\textbf{79.74}&\cellcolor{gray!20}\textbf{94.82}&\cellcolor{gray!20}\textbf{68.33} \\\midrule
        \multirow{2}{*}{LaBO~\cite{labo}}&\multirow{2}{*}{LLM}   &baseline &81.91 &56.24&97.14 &84.15&97.41 &74.56&99.90 &63.17\\
        &  &\cellcolor{gray!20}\textbf{+DDO} &\cellcolor{gray!20}\textbf{82.34}&\cellcolor{gray!20}\textbf{59.60} &\cellcolor{gray!20}\textbf{97.26} &\cellcolor{gray!20}\textbf{87.66}&\cellcolor{gray!20}\textbf{98.12} &\cellcolor{gray!20}\textbf{80.00} &\cellcolor{gray!20}\textbf{99.93} &\cellcolor{gray!20}\textbf{68.01} \\
        \bottomrule[1pt]
    \end{tabular}}
\caption{Detailed accuracy performance comparison on single unseen domain benchmarks, including CUB-Painting, AwA2-clipart, LADA-Sculpture, and LADV-3D.}
\label{Table:singleunseenAppendix}
\end{table*}

\begin{table*}[tp!]
\centering
\resizebox{0.75\textwidth}{!}{
    \begin{tabular}{l|l|c|cccccc}
        \toprule[1pt]
        \multicolumn{9}{c}{DomainNet.} \\\midrule
        &&ID&\multicolumn{6}{c}{OOD}\\
        Model &Method &real&clipart &infograph &painting &quickdraw &Sketch & Avg  \\\midrule
        \multicolumn{9}{c}{CLIP ViT-B/32} \\\midrule
        \multirow{2}{*}{LaBO~\cite{labo}} &baseline &86.11 & 60.60 &35.00 & 54.64 &8.40 &49.7 &41.67   \\
        &\cellcolor{gray!20}\textbf{+DDO} &\cellcolor{gray!20}\textbf{86.33} &\cellcolor{gray!20} \textbf{64.00} &\cellcolor{gray!20}\textbf{39.66} &\cellcolor{gray!20} \textbf{58.90} &\cellcolor{gray!20}\textbf{8.57} &\cellcolor{gray!20}\textbf{52.9}&\cellcolor{gray!20}\textbf{44.81}\\\midrule
        \multicolumn{9}{c}{CLIP ViT-L/14} \\\midrule
        \multirow{2}{*}{LaBO~\cite{labo}} &baseline &91.20 &76.04 &48.41 &66.16 &16.58 &66.35 &55.63  \\
        &\cellcolor{gray!20}\textbf{+DDO} &\cellcolor{gray!20}\textbf{91.29} &\cellcolor{gray!20} \textbf{77.37} &\cellcolor{gray!20}\textbf{53.00} &\cellcolor{gray!20} \textbf{68.91} &\cellcolor{gray!20}\textbf{17.27} &\cellcolor{gray!20}\textbf{69.04} &\cellcolor{gray!20}\textbf{56.20}\\
        \bottomrule[1pt]
    \end{tabular}}
\caption{Detailed accuracy performance comparison on multiple unseen domain benchmarks, $\ie$ DomainNet.}
\label{Table:multipleunseendomainAppendix}
\end{table*}

\begin{table*}[thb!]
\centering
\resizebox{0.65\textwidth}{!}{
    \begin{tabular}{l|l}
        \toprule[1pt]
        Dataset & Relevant descriptors (keywords) \\\midrule
        CUB-Painting & 
        \begin{tabular}[t]{@{}l@{}}
        painting, sketch, watercolor, drawing, \\ 
        doodle, art
        \end{tabular} \\\midrule
        AwA2-clipart & 
        \begin{tabular}[t]{@{}l@{}}
        clipart, cartoon, emoji, comic, anime, \\ 
        avatar, animated
        \end{tabular} \\\midrule
        LADA-Sculpture & 
        \begin{tabular}[t]{@{}l@{}}
        sculpture, 3D, statue
        \end{tabular} \\\midrule
        LADV-3D & 
        \begin{tabular}[t]{@{}l@{}}
        3D, CGI, VR, low-poly
        \end{tabular} \\\midrule
        DomainNet & 
        \begin{tabular}[t]{@{}l@{}}
        painting, clipart, infographic, quickdraw, \\ 
        sketch, watercolor, cartoon, collage, art, \\ 
        drawing, sketch, illustration, doodle, poster \\
         emoji, comic, anime \\
        \end{tabular} \\\midrule
        \bottomrule[1pt]
    \end{tabular}}

\caption{Relevant descriptors for each benchmark.}
\label{Table:relevant}
\end{table*}

\begin{table*}[tp!]
\centering
\resizebox{0.75\textwidth}{!}{
    \begin{tabular}{l|c|c|c|c|c|c|c}
        \toprule[1pt]
        & &real &clipart &painting &infograph &sketch &quickdraw\\\midrule
        \multirow{3}{*}{LanCE}   &baseline  &91.20 &76.04 &66.16 &48.41&66.35&16.58\\
        &\textbf{+DDO(IR)} &91.20&76.60&67.80&50.00&67.74&16.30 \\
        &\textbf{+DDO} &\textbf{91.29}  &\textbf{77.37} &\textbf{68.91} &\textbf{53.0}&\textbf{69.04}&\textbf{17.27} \\
        \bottomrule[1pt]
    \end{tabular}}
\caption{Ablation studies on DomainNet for the effect of relevance of the domain descriptors. +DDO(IR) only use the domain-irrlevant descriptors while +DDO use all domain descriptors.}
\label{Table:abforremoveingDomainNet}
\end{table*}

\section{More qualitative results}
\label{appendix:qualitative}
Fig.~\ref{qualitative} has shown more qualitative results about the top-5 visual concepts, ranked by the weights in $W_F$, demonstrating that our proposed DDO regularizer can reduce the correlation between domain-specific concepts and final predictions.

\section{Human evaluation}
\label{appendix:human}

To validate the efficacy of our proposed method, we conduct a human evaluation on the top 10 visual concepts that exhibit a high correlation with the final class, ranked by the weights $W_F$ trained on DomainNet. 
The evaluation considers two key aspects: \textit{Discriminability} and \textit{Generalizability}. 
For each concept, we present several images from all visual domains and invite three human experts to assign a score ranging from 0 to 4.
Specifically, for \textit{Discriminability}, a score of 0 indicates the concept is unrelated to the corresponding category, while a score of 4 signifies the concept is a salient visual feature for the category. For \textit{Generalizability}, a score of 0 indicates the concept exists only in a single domain, whereas a score of 4 represents a domain-invariant concept. The scores from the three annotators are averaged, and concepts with an average score greater than 2 are classified as either discriminative or domain-invariant concepts. For each concept, we generate a binary label based on these classifications. Finally, we analyze the percentage of discriminative and domain-invariant concepts to report the final results.
To evaluate these two metrics, we select top-10 concepts for each class ranking by their weights in the final layer $W_F$, and ask annotators to judge whether each concept meets the demands above. The ratio of accurate concepts are shown in the Table.~\ref{Table:human} where our proposed LanCE achieves better results than the baseline LaBO, demonstrating the effectiveness of languid-guided concept erasing design can significantly decrease the association between domain-specific concepts and the final output.

\section{Limitations} 
Our method highly depends on pre-trained VLMs like CLIP and LLMs like GPT-3.5. However, these models are limited in application to some professional fields like medical treatments. We think further integration of an extra knowledge base and task-specific fine-tuning of these pre-trained models is a potential solution to solve these limitations. We hope this work can prompt the development of robust interpretable models.

\begin{figure*}[!thb]
	\centering 
	\includegraphics[width=1.0\textwidth]{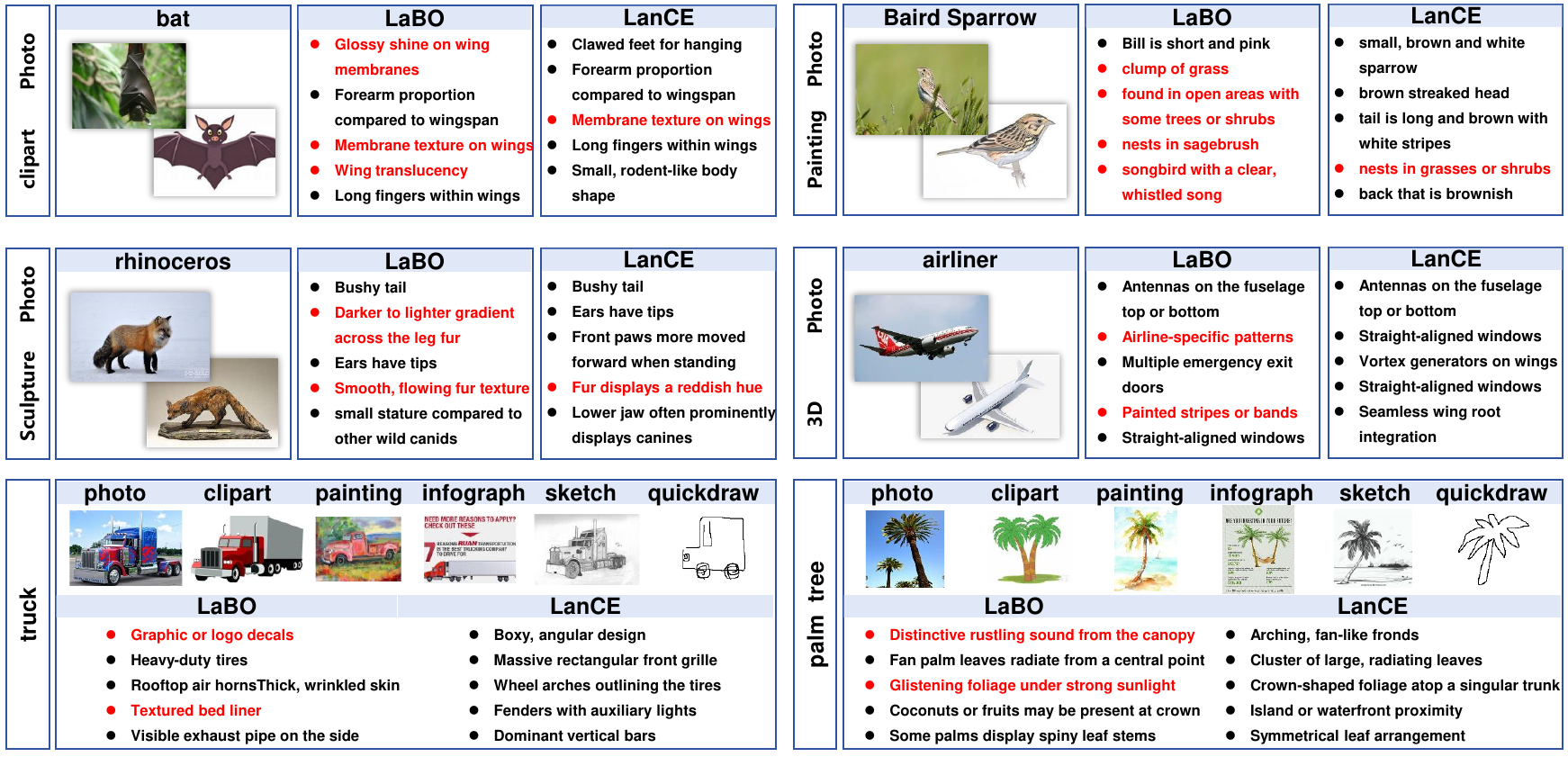}
	\caption{More qualitative results. top-5 concepts, ranking by their weights in the final linear layer WF . Baseline indicates the
results of the original LaBo. Domain-specific concepts are hightlighted in red.}\label{qualitative}
\end{figure*}

\begin{table}[tp!]
\centering
\resizebox{0.50\textwidth}{!}{
    \begin{tabular}{l|c|c}
        \toprule[1pt]
        Model&Discriminability($\%$) &Generalizability($\%$) \\\midrule
        baseline  &75 &64 \\
        \textbf{LanCE} &\textbf{79} &\textbf{82} \\
        \bottomrule[1pt]
    \end{tabular}}
    \vspace{-3mm}
\caption{Human evaluation about the percentage of distinguishing concepts and percentage of domain-invariant concepts.}
\label{Table:human}
    \vspace{-2mm}
\end{table}

\end{document}